\documentclass[sigconf]{acmart}

\AtBeginDocument{%
  }

\setcopyright{rightsretained}
\copyrightyear{2024}
\acmYear{2024}
\acmDOI{10.1145/3641517.3664388}

\acmConference{SIGGRAPH Emerging Technologies '24}{July 27 - August 01, 2024}{Denver, CO, USA}

\acmBooktitle{Special Interest Group on Computer Graphics and Interactive Techniques Conference Emerging Technologies (SIGGRAPH Emerging Technologies '24), July 27 - August 01, 2024}
\acmISBN{979-8-4007-0524-3/24/07}

\begin{document}

\title{Controlling the color appearance of objects by optimizing the illumination spectrum}

\author{Mariko Yamaguchi}
\affiliation{%
  \institution{NTT Corporation}
  \country{Japan}
}
\email{mrk.yamaguchi@ntt.com}

\author{Masaru Tsuchida}
\affiliation{%
  \institution{NTT Corporation}
  \country{Japan}
}
\email{masaru.tsuchida@ntt.com}

\author{Takahiro Matsumoto}
\affiliation{%
  \institution{NTT Corporation}
  \country{Japan}
}
\email{takahiro.matsumoto@ntt.com}

\author{Tetsuro Tokunaga}
\affiliation{%
  \institution{NTT Corporation}
  \country{Japan}
}
\email{tetsuro.tokunaga@ntt.com}

\author{Takayoshi Mochizuki}
\affiliation{%
  \institution{NTT Corporation}
  \country{Japan}
}
\email{takayoshi.mochiduki@ntt.com}

\renewcommand{\shortauthors}{Mariko et al.}

\begin{abstract}
We have developed an innovative lighting system that changes specific target colors while keeping the lights appearing naturally white. 
By precisely controlling the spectral power distribution (SPD) of illumination and harnessing the unique phenomenon of metamerism, our system achieves unique color variations in ways you’ve never seen before.
Our system calculates the optimal SPDs of illumination for given materials to intensively induce metamerism, and then synthesizes the illumination using various colors of LEDs.
We successfully demonstrated the system’s implementation at Paris Fashion Week 2024. As models step onto the stage, their dresses initiate a captivating transformation.
Our system altering the colors of the dresses, showcasing an impressive transition from one stunning color to another.
\end{abstract}



\begin{teaserfigure}
  \centering
  \vspace{-0.1cm}
  \includegraphics[height=3.7cm]{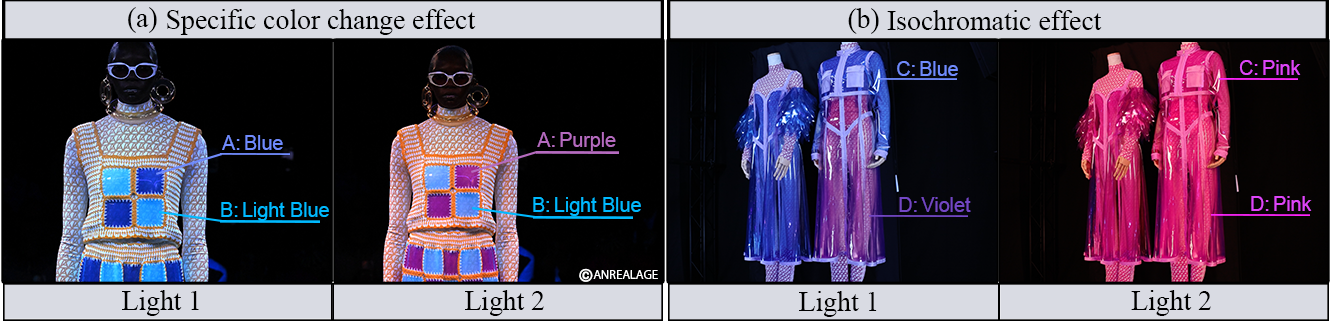}
  \vspace{-0.1cm}
  \caption{Our system has two technical contributions, which are visual effects that change the color appearance of objects by controlling the distribution of the illumination spectrum. (a) Specific color change effect: Comparing the color appearance of the dress under Light 1 and Light 2, the color of material A changes significantly, while material B remains its color, even after switching from Light 1 to Light 2. (b) Isochromatic effect: Materials C and D appear to be different colors under Light 1, while they appear to be the same color under Light 2. In both cases, we achieved designing Light 1 and Light 2 as a natural white appearance.}
  \label{fig1}
\end{teaserfigure}


\maketitle

\vspace{-0.1cm}
\section{Introduction}
Although the colors of materials typically remain consistent under one light source, they may appear different under another light source.
This phenomenon, known as metamerism, presents challenges for designers and color managers. 
Metamerism also occurs inadvertently in everyday life, manifesting as minor color variations. 
Most of the previous research on metamerism has focused on ways of diminishing its adverse effect \cite{David2020}.

In contrast, our research aims to strategically utilize this phenomenon by manipulating lighting to achieve desired color perceptions, thereby creating a mesmerizing visual impact. 
We propose a novel lighting system that synthesizes the SPD of an illumination optimized for a material to induce metamerism. 
Our goal is to create two visual effects using our lighting system. 
First, the isochromatic effect makes two objects of different colors appear to be the same color. 
Second, the specific color change effect changes the only one color of objects. 
To achieve this goal, our system determines the optimal combination of illumination for the given materials. 
It can identify combinations of materials with spectral reflectance under two distinct illuminations, and alternatively, it can determine the optimal combination of two illuminations for the given materials.
In this paper, we describe the latter approach.

\vspace{-0.1cm}
\section{METHODS}
The CIE XYZ value $\boldsymbol{v} = \boldsymbol{[X, Y, Z]^T}$ of the object under an illumination with the SPD $\boldsymbol{w}$ is represented as a matrix formulation $\boldsymbol{v} = \boldsymbol{CRw}$ \cite{TsuchidaKH21}.
In this formulation, $\boldsymbol{R}$ is the diagonal matrix of the object spectral reflectance and $\boldsymbol{C}$ is the CIE color matching function $\boldsymbol{C} =\boldsymbol{[C_X,C_Y,C_z]^T}$.
Metamerism is a perceived matching of colors with different SPDs, and it can be represented mathematically as $\boldsymbol{CR_1 w}=\boldsymbol{CR_2 w}$ and $\boldsymbol{CR_1 w'} \neq \boldsymbol{CR_2 w'}$.

Utilizing the above theory of object color appearance, we achieve the isochromatic effect, the proposed system identifies combinations of $\boldsymbol{w_1}$, and $\boldsymbol{w_2}$ that satisfy Eq. (\ref{eq1}).

\begin{equation}
\begin{split}
&\underset{\boldsymbol{w_1}}{maximize} \;\;\; ||U(\boldsymbol{CR_1 w_1}) - U(\boldsymbol{CR_2 w_1})||, \\
s.t. \;\;\; &||U(\boldsymbol{C R_2 w_2}) - U(\boldsymbol{C R_2 w_1})|| \le \delta, \\ 
&||U(\boldsymbol{C w_1}) - U(\boldsymbol{C w_2})|| \le \delta_{white}, \; |\boldsymbol{C_Y w_1} - \boldsymbol{C_Y w_2}| \le \delta_{Y}
\label{eq1}
\end{split}
\end{equation}

$U()$ is an operator for calculating CIE $u'v'$ values of a given spectrum and $|| \boldsymbol{U} ||$ is an operator for calculating $\{{(u')}^2+{(v')}^2\}^{1/2}$, where $\boldsymbol{U}$ is the value on the CIE $u'v'$ coordinates. 
We set $\delta_{Y}$, representing the difference of brightness between two illuminations, to 0. 
$\delta$ and $\delta_{white}$ are parameters tuned by the designer.
These parameters affect the impression created by changing the color of target objects and the naturalness of the white appearance when comparing the lighting before and after the switch. 
To achieve the specific color change effect, the first two terms in Eq. (\ref{eq1}) are modified as follows: $\underset{\boldsymbol{w_1, w_2}}{maximize}||U(\boldsymbol{CR_1 w_1}) - U(\boldsymbol{CR_1 w_2})||$ and $||U(\boldsymbol{C R_2 w_1}) - U(\boldsymbol{C R_2 w_2})|| \le \delta$. 

We modeled the SPD of the illumination $\boldsymbol{w_i} (i=1, 2)$ as a linear combination of LEDs. $\boldsymbol{w_i}$ is represented using the SPD of LEDs as $\boldsymbol{w_i} = \sum_{(k=1)}^{N}\alpha_{i,k} \boldsymbol{e}_k$, where $\alpha_{i,k}$ and $N$ represent the intensity parameters of each LED and the number of LEDs, respectively. 
$\boldsymbol{e}_k$ takes different values for each LED. 
$\boldsymbol{w_1}$ and $\boldsymbol{w_2}$ are determined by finding $\alpha_{1, k}$ and $\alpha_{2, k}$ for all LEDs.
This simplifies the complex calculation and enables us to focus on adjusting the LED weights to achieve desired outcome.

\vspace{-0.1cm}
\section{IMPLEMENTATION AND RESULTS}

We demonstrated the system’s implementation at Paris Fashion Week 2024. 
We utilized eight sets of fifteen-channel visible LEDs (LEDCube, Thouslite Inc.) and twelve UV lights for the show illumination. 
The proposed system controlled the individual brightness of lights via PC with synchronization. 
All lights were positioned from truss frames in a box shape to centrally illuminate the stage. 
We determined the parameters in Eq. (\ref{eq1}) as $\delta = 1.0 * 10^{-1}$ and $\delta_{white} = 8.5 * 10^{-2}$ after discussions with the designer.
The dresses used in our demonstration are made from several types of photochromic materials to enhance visual impressions.
The photochromic material changes from clear to colored when exposed to UV light. As a model steps onto the stage, UV lights activate, revealing the colors
of the photochromic materials. 
Our system then enhances and changes the colors of the dress, showcasing an impressive transition from one stunning color to another. 
In the demonstration of the specific color change effect, the blue and light-blue textures of the dress transform into a combination of purple and light-blue, while preserving the appearance of light-blue texture and white illumination (Fig. \ref{fig1}(a)). 
In the demonstration of the isochromatic effect, the blue and violet textures of the two dresses transform into a pink texture only, while preserving the appearance of white illumination (Fig. \ref{fig1}(b)).

The demonstrations were successfully presented, changing the target colors of the dress and creating an impressive experience for the audience.
We measured the color variations of the synthesized illumination and materials using a spectroradiometer under the same condition as the demonstrations. Color variations were evaluated in CIE $u'v'$ color space. 
Regarding the isochromatic effect, the color variation of two materials with spectral reflectance $\boldsymbol{R_1}$ and $\boldsymbol{R_2}$ under illuminations $\boldsymbol{w_1}$ was $1.9 * 10^{-1}$, and under illumination $\boldsymbol{w_2}$, it was $9.8 * 10^{-2}$.
The color variation of white due to the shift from illumination $\boldsymbol{w_1}$ to $\boldsymbol{w_2}$ was $8.1 * 10^{-2}$.
In terms of the specific color change effect, the color variations of the material with spectral reflectance $\boldsymbol{R_1}$ when switching from illumination $\boldsymbol{w_1}$ to $\boldsymbol{w_2}$ was $4.8 * 10^{-2}$.
For the material with spectral reflectance $\boldsymbol{R_2}$, the variation was $3.6 * 10^{-3}$.
The color variation of white due to the shift from illumination $\boldsymbol{w_1}$ to $\boldsymbol{w_2}$ was $3.7 * 10^{-2}$.
These results demonstrate that the lighting controls determined with the parameters $\delta$ and $\delta_{white}$ changed the material colors in real environments, as predicted by the proposed method. 
This shows that we succeeded in inducing both metamerism effects intensively. 
Furthermore, we intensively induced two types of metamerism effects and created a mesmerizing visual impact with our lighting system.

\vspace{-0.1cm}
\section{CONCLUSION}
We have developed a lighting system that changes specific target colors while maintaining a natural white appearance, using precise spectral control of the SPD of illumination and metamerism.
It was successfully showcased at a fashion show in Europe. Our system’s capabilities extend beyond show business, with potential applications in industrial design, advertising, inspection, and repair across fields such as apparel, home electronics, automotive, and architecture.

\vspace{-0.1cm}
\begin{acks}
ANREALAGE CO., LTD. and IMAGICA EEX Inc. assisted in the creation of the stage performance and costumes.
\end{acks}

\vspace{-0.1cm}
\bibliographystyle{ACM-Reference-Format}
\bibliography{sample-base}


\end{document}